\documentclass[twoside]{article}

\usepackage[accepted]{aistats2025}
\makeatletter
\renewcommand{\AISTATS@appearing}{}
\makeatother
\makeatletter
\renewcommand{\@copyrightspace}{}
\makeatother

\usepackage[utf8]{inputenc} %
\usepackage[T1]{fontenc}    %
\usepackage{hyperref}       %
\usepackage{url}            %
\usepackage{booktabs}       %
\usepackage{amsfonts}       %
\usepackage{nicefrac}       %
\usepackage{microtype}      %
\usepackage{xcolor}         %

\usepackage{mathtools}
\usepackage{amsmath}
\usepackage{amssymb}
\usepackage{amsthm}
\theoremstyle{plain} %
\newtheorem{definition}{Definition}

\theoremstyle{plain}
\newtheorem{theorem}{Theorem}
\newtheorem{lemma}{Lemma}
\newtheorem{corollary}{Corollary}
\newtheorem{proposition}{Proposition}

\usepackage{xcolor}
\usepackage{graphicx}
\usepackage{subfigure}
\usepackage{enumitem}
\usepackage{booktabs}

\usepackage{multirow}

\usepackage{setspace}
\usepackage{marginnote}

\usepackage{breqn}
\usepackage[round]{natbib}
\renewcommand\cite[1]{\citep{#1}}

\DeclarePairedDelimiter\multiset{\lbrace\!\!\lbrace}{\rbrace\!\!\rbrace}

\DeclareMathOperator{\Hash}{\text{\textsc{Hash}}}

\DeclareMathOperator{\Readout}{\text{\textsc{ReadOut}}}

\DeclareMathOperator{\MLP}{\mathrm{MLP}}
\DeclareMathOperator{\ReLU}{\mathrm{ReLU}}

\DeclareMathOperator*{\argmax}{arg\,max} %

\newcommand{\LK}[1]{\textcolor{black}{#1}}%

\begin{document}

\runningtitle{On the Relationship Between Robustness and Expressivity of Graph Neural Networks}

\runningauthor{Lorenz Kummer, Wilfried N.~Gansterer,  Nils M.~Kriege}

\twocolumn[

\aistatstitle{On the Relationship Between Robustness \\ and Expressivity of Graph Neural Networks}

\aistatsauthor{ Lorenz Kummer\textsuperscript{1,2} \And Wilfried N. Gansterer\textsuperscript{1} \And  Nils M.~Kriege\textsuperscript{1,3} }

\aistatsaddress{ 
\textsuperscript{1}Faculty of Computer Science \\
University of Vienna \\ 
Vienna, Austria
\And
\textsuperscript{2}Doctoral School Computer Science \\ 
University of Vienna \\ 
Vienna, Austria 
\And 
\textsuperscript{3}Research Network Data Science \\
University of Vienna \\ 
Vienna, Austria} ]
\begin{abstract}
  We investigate the vulnerability of Graph Neural Networks (GNNs) to bit-flip attacks (BFAs) by introducing an analytical framework to study the influence of architectural features, graph properties, and their interaction.
  The expressivity of GNNs refers to their ability to distinguish non-isomorphic graphs and depends on the encoding of node neighborhoods. We examine the vulnerability of neural multiset functions commonly used for this purpose and establish formal criteria to characterize a GNN's susceptibility to losing expressivity due to BFAs. This enables an analysis of the impact of homophily, graph structural variety, feature encoding, and activation functions on GNN robustness. We derive theoretical bounds for the number of bit flips required to degrade GNN expressivity on a dataset, identifying ReLU-activated GNNs operating on highly homophilous graphs with low-dimensional or one-hot encoded features as particularly susceptible. Empirical results using ten real-world datasets confirm the statistical significance of our key theoretical insights and offer actionable results to mitigate BFA risks in expressivity-critical applications.
\end{abstract}

\section{INTRODUCTION} 
\label{sec:intro}
Graph Neural Networks (GNNs) have recently emerged as a dominant technique for machine learning on structured data. Complex objects such as molecules, proteins, or social networks can be represented as graphs, with nodes as components and edges as their relations, both potentially carrying features. By generalizing deep learning techniques to graphs, GNNs open up new domains, including the analysis of financial and social networks, medical data, as well as chem- and bioinformatics~\cite{lu2021weighted, %
sun2021disease, gao2022cancer, wu2018moleculenet, xiong2021novo}. Their increasing adoption across various fields necessitates examining robustness and potential security issues. Traditionally, adversarial attacks on GNNs target input graph data~\cite{wu2022graph}, aiming to induce perturbations before training (\emph{poisoning}) to produce faulty models~\cite{ma2020towards, wu2022graph}, or during inference (\emph{evasion attacks}) to degrade prediction quality. These perturbations are achieved by modifying node features, adding or deleting edges, or injecting new nodes~\cite{sun2019node, wu2022graph}. An overview and classification of existing attacks and corresponding defenses, along with representative algorithms, is provided by \citet{jin2021adversarial}.

A novel paradigm for attacking GNNs is bit-flip attacks (BFAs). In contrast to the orthogonal approaches of poisoning and evasion, BFAs %
degrade the model directly by manipulating the binary in-memory representations of the model's trainable parameters.
The feasibility of BFAs, which are typically executed at inference time, 
is well-documented~\cite{
mutlu2019rowhammer, 
yao2020deephammer, breier2018practical}.
However, unlike Convolutional Neural Networks (CNNs), where the security issues of BFAs are thoroughly studied~\cite{qian2023survey, khare20222design} and largely attributed to weight sharing in convolutional filters~\cite{hector2022evaluating}, the robustness of GNNs, which lack such filters%
, remains underexplored, despite their use in safety-critical domains such as medical diagnoses~\cite{gao2022cancer, lu2021weighted}, %
electronic health record modeling~\cite{sun2021disease}%
, and drug development~\cite{xiong2021novo}. %
To date, random~\cite{jiao2022assessing} and gradient-based~\cite{wu2023securing, kummer2024attacking} BFAs on GNNs have primarily been studied empirically. The intricate mechanisms determining GNN vulnerability and their relation to learning tasks, graph properties and key properties such as expressivity remain poorly understood theoretically. Expressivity in particular is a fundamental factor, even if in practice, other factors that go beyond expressivity influence predictive performance as well. Any GNN solving a certain node classification task must at least pairwise distinguish those nodes associated with different classes. This equates to such nodes being mapped to embedding vectors of which at least one element is different, reflecting (for a message passing based GNN with $k$ layers) that the two nodes’ unfolding trees of height $k$ are non-isomorphic. Likewise, any model solving a certain graph classification task must at least pairwise distinguish those non-isomorphic graphs belonging to different classes, implying that for each pair the set of node embeddings must be different in at least one element, linking again to the unfolding tree isomorphism problem and therefore expressivity.

\paragraph{Related Work}
Research on GNN resilience to bit flips is scarce: \citet{jiao2022assessing} studied random bit faults in floating-point hardware and observed that resilience varies significantly across models and datasets, but without providing a theoretical explanation. Focusing on detecting bit flips in GNN weights,~\citet{wu2023securing} transferred the seminal BFA of \citet{rakin2019bitflip} from quantized CNNs to GNNs, modifying it to flip bits in the exponent of a weight's floating-point representation. \citet{amir2024neural}'s results indicate that GNN expressivity is generally resilient to random bit flips. Thus, \citet{kummer2024attacking} proposed a BFA targeting quantized GNNs, exploiting their expressivity to degrade classification accuracy by targeting injectivity in neighborhood aggregation. %
Given the key role of expressivity in GNNs, it is crucial to better understand its relationship to robustness against BFAs.

\paragraph{Contribution} 
We characterize the properties of bit flips degrading GNN expressivity and establish formal bounds on the number of bit flips required. We first analyze general vulnerabilities of neural multiset functions based on element-wise summation and then focus on derived GNNs, where instances like Graph Isomorphism Network (GIN) achieve an expressivity equivalent to the 
Weisfeiler-Leman test~\cite{leskovec2019powerful}.
We show that a GNN's vulnerability to expressivity degradation from bit flips depends on parameter count, number of bits used to represent a value (bit width), diversity of distinguishable subgraphs, variety of GNN embeddings, input feature encoding, dataset homophily, activation function, and the bit's location in the numerical representation.

\subsection{Preliminaries}
A \emph{graph} $G$ is a pair $(V,E)$ of a finite set of \emph{nodes} $V$ and \emph{edges} $E \subseteq \left \{ \left \{u,v \right \} \subseteq V \right \}$. 
The set of nodes and edges of $G$ is denoted by $V(G)$ and $E(G)$, respectively. The \emph{neighborhood} of $v$ in $V(G)$ is $N(v) = \left \{ u \in V(G) \mid \{u, v\} \in E(G) \right \}$.
If there exists a bijection $\varphi \colon V(G) \rightarrow V(H)$ with $\{u,v\}$ in $E(G)$ if and only if $\{\varphi(u),  \varphi(v)\}$ in $E(H)$ for all $u$, $v$ in $V(G)$, we call the two graphs $G$ and $H$ \emph{isomorphic} and write $G \simeq H$. For two graphs with roots $r\in V(G)$ and $r'\in V(H)$, the bijection must additionally satisfy $\varphi(r)=r'$. The equivalence classes induced by $\simeq$ are referred to as \emph{isomorphism types}. 
A \emph{node colored} or \emph{labeled graph} is a pair $(G,l)$ consisting of a graph $G$ endowed with a \emph{node coloring} $l\colon V(G) \rightarrow \Sigma$.
We call $l(v)$ a \emph{label} or \emph{color} of $v \in V(G)$. 
We denote a multiset by $\multiset{\dots}$.

\paragraph{The Weisfeiler-Leman Algorithm}
\label{ssec:1wl}
Let $(G,l)$ denote a labelled graph. In every iteration $t > 0$, a node coloring $c_l^{(t)} \colon V(G) \rightarrow \Sigma$ is computed, which depends on the coloring $c_l^{(t-1)}$ of the previous iteration. At the beginning, the coloring is initialized as $c_l^{(0)} = l$. In subsequent iterations $t > 0$, the coloring is updated according to
    $c_l^{(t)}(v) = \Hash\left ( c_l^{(t-1)}(v), \multiset{c_l^{(t-1)}(u)|u \in N(v)} \right ),$
  where $\Hash$ is an injective mapping of the above pair to a unique value in $\Sigma$, that has not been used in previous iterations. 
The $\Hash$ function can, for example, be realized by assigning new consecutive integer values to pairs when they occur for the first time~\cite{shervashidze2011weisfeiler}.
Let $C_l^{(t)}(G)=\multiset{c_l^{(t)}(v) \mid v \in V(G)}$ be the multiset of colors a graph exhibits in iteration $t$.
The iterative coloring terminates if $|C_l^{(t-1)}(G)|=|C_l^{(t)}(G)|$, i.e., the number of colors does not change between two iterations. 
For testing whether two graphs $G$ and $H$ are isomorphic, the above algorithm is run in parallel on both $G$ and $H$.
If $C_l^{(t)}(G) \neq C_l^{(t)}(H)$ for any $t \geq 0$, then $G$ and $H$ are not isomorphic. 
The label $c_l^{(t)}(v)$ assigned to a node $v$ in the $t$th iteration of the 1-WL test encodes the isomorphism type of the tree of height $t$ representing the $t$-hop neighborhood of $v$, see%
~\cite{dinverno2021aup,Jegelka2022GNNtheory, schulz2022weisfeiler} for details.

\paragraph{From DeepSets to Graph Neural Networks}
DeepSets~\cite{ZaheerKRPSS17} is an architecture for modeling set functions, suitable for tasks with unordered data. Given a set \( X = \{\mathbf{x}_1, \mathbf{x}_2, \dots, \mathbf{x}_n\} \), a deep set function is of the form \( f(X) = \rho\left(\sum_{i=1}^k \phi(\mathbf{x}_i) \right) \). Due to the summation, $f$ is invariant to the permutation of the input elements, capturing the intrinsic property of sets. The function \( \phi \) maps each $\mathbf{x}_i \in X$ to a feature space, in which their sum is calculated and then transformed by \( \rho \) into the final output. Typically both functions are realized by a Multi-Layer Perceptron (MLP), whose universal approximation property leads to universal approximators of set functions~\cite{ZaheerKRPSS17}, albeit with certain practical caveats~\cite{WagstaffFEOP22}. This concept extends to multisets, reflecting both identity and multiplicity of elements~\cite{leskovec2019powerful,amir2024neural}.

In this context, GNNs can be seen as stacked neural (multi)set functions, where each layer aggregates and combines node features to reflect the multiset nature of graph neighborhoods, a technique also known as \emph{message-passing} (MP). Specifically, a \emph{neural moment} function \( \Hat{f} \) derived from sum-pooling operations is given by
$\Hat{f}(\multiset{\mathbf{x}_1, \dots, \mathbf{x}_k}) = \sum_{i=1}^k f(\mathbf{x}_i)$, 
where \( f\colon V^d \to V^m \) is an MLP mapping elements from a set \( V^d \) to a vector space \( V^m \)~\cite{amir2024neural}. This ensures that for a suitable choice of $d$ and $m$ and parameterization of the MLP, each multiset is uniquely represented by its moment, allowing distinct multisets to have distinct representations, as in DeepSets.
GNNs apply this principle by using an MLP in each layer's update rule,
\begin{equation}
    \mathbf{h}_v^{(k)} = \MLP^{(k)}\left(\mathbf{h}_v^{(k-1)} + \sum_{u \in N(v)} \mathbf{h}_u^{(k-1)}\right),
    \label{eq:defginsimple}
\end{equation}
where \( \mathbf{h}_v^{(k)} \) represents the embedding of node \( v \) at layer \( k \).
This effectively extends DeepSets to graph-structured data, as the layer-wise combination of node features from a node and its neighborhood allows such a model to learn to distinguish certain graph structures.
If the input features are one-hot encodings, this ensures injectivity of the summation at the first layer, even without an initial MLP. For graph-level tasks, the readout function typically aggregates the outputs across all layers to obtain a graph-level embedding $\mathbf{h}_G$, where \( \Vert \) denotes concatenation: 
\begin{equation}
    \mathbf{h}_G = \Big\Vert_{k=0}^n \sum_{v \in V(G)} \mathbf{h}_v^{(k)}.
    \label{eq:defginreadout}
\end{equation}
The update rule can also be expressed in matrix form using the adjacency matrix \( \mathbf{A} \) and node feature matrix %
    $\mathbf{H}^{(k)} = \MLP^{(k)}\left((\mathbf{A} + \mathbf{I}) \cdot \mathbf{H}^{(k-1)}\right)$,
which allows for an easier notation of a GNN as $\Phi^{(k)}\colon {D} \rightarrow {Q}^{m_k}$, mapping domain ${D} = \{ (\mathbf{A}_1, \mathbf{X}_1), \dots (\mathbf{A}_n, \mathbf{X}_n) \}$ to some codomain ${Q}^{m_k}$ with $\mathbf{H}^{(0)} = \mathbf{X}$. 
For node features $\mathbf{X}$ representing encodings of the graphs labels, the 
set ${D} = \{ (\mathbf{A}_1, \mathbf{X}_1), \dots (\mathbf{A}_n, \mathbf{X}_n) \}$ is only a different notation for the set of labeled graphs $\{ (G_1, l_1), \dots  (G_n, l_n)\}$. Hence, we do not distinguish between the two in the following,  
as it will be clear from the context which one is used.

Graph Isomorphism Network (GIN)~\cite{leskovec2019powerful} extends this approach, achieving expressivity provably equal to 1-WL. GIN's update rule differs from Equation~\eqref{eq:defginsimple} (and its matrix notation) by multiplying the embeddings $\mathbf{h}_v^{(k-1)}$ by a learnable parameter $(1+\epsilon^{(k)})$, differentiating a node's own features from aggregated neighbor features. While much research aims to surpass the expressivity of GIN and the 1-WL test~\cite{wlsurvey}, neighborhood aggregation remains prevalent, and 1-WL distinguishes most graphs in common benchmarks~\cite{Zopf22a, morris2021power}.

\section{BIT FLIPS AND EXPRESSIVITY}
\label{sec:bitflips_expressivty}
We first examine the general properties and vulnerabilities of neural moments, followed by a detailed analysis of neural moment-%
based GNN architectures. %

\label{ssec:neumom}
Let $\Omega_n^d$ be the set of all bounded multisets with at most $n \in \mathbb{N}_{\geq 1}$ elements from some 
$d \in \mathbb{N}_{\geq 2}$ dimensional vector space $V^d$. For an $X \in \Omega_n^d$, we denote as $m_X(\mathbf{x})$ the multiplicity function returning the number of occurrences of $\mathbf{x} \in X$ and as $S(X)$ the support set of $X$, the set of $X$'s distinct elements. Further, let the moment $\Hat{f} : \Omega_n^d \rightarrow V^m$ induced by some function $f : V^d \rightarrow V^m$ be defined as $\Hat{f}(X) = \sum_{\mathbf{x} \in X} f(\mathbf{x})$~\cite{amir2024neural}.
We now explore the relationship between the pre-aggregation linearly independent mapping of distinct multiset elements by $f$, the injectivity of $f$ and moment injectivity of $\Hat{f}$.

A sufficient condition for \( f \) to be injective is that for any multiset \( A \in \Omega_n^d \), the vectors $\multiset{f(\mathbf{x}) \mid \mathbf{x} \in S(A)}$ are linearly independent.
Next, we show that for the set of bounded multisets $\Omega_n^d$ with $n \geq 2$, a pre-aggregation linearly independent mapping of distinct multiset elements by $f$ implies moment injectivity (i.e. the injective mapping of distinct multisets) of $\Hat{f}$. We chose $n \geq 2$ because for $n = 1$, the linear independence of the pre-aggregation mapping of distinct multiset elements is always trivially given since each multiset has at most one element.
\proposition[Sufficient condition for moment injectivity]{
\label{lem:mulinjlu}
Let \( \Hat{f}: \Omega_n^d \rightarrow V^m \) denote the moment induced by any \( f: V^d \rightarrow V^m \). Then for \( n \geq 2 \), the second statement follows from the first:

1. For all \( A \in \Omega_n^d \), if 
\(
\sum_{\mathbf{x} \in S(A)} f(\mathbf{x}) \lambda_{\mathbf{x}} = \mathbf{0}, \quad \lambda_{\mathbf{x}} \in V,
\)
then \( \lambda_{\mathbf{x}} = 0 \) for all \( \mathbf{x} \in A \).
\\
2. For all \( A, B \in \Omega_n^d \) \( \Hat{f}(A) = \Hat{f}(B) \) implies \( A = B \).

}\normalfont

To summarize, non-injectivity of $f$ implies, in particular, linear dependence of the elements $\multiset{f(\mathbf{x}), \mathbf{x} \in A}{}$. Proposition~\ref{lem:mulinjlu} guarantees moment injectivity of $\hat{f}$ only under linear independence. Thus, without linear independence, moment injectivity of $\hat{f}$ is not guaranteed, and we cannot exclude the possibility that $\hat{f}$ fails to distinguish distinct multisets.
This is consistent with %
the finding that projection into a low rank sub-space is the underlying cause of indistinguishable node embeddings in GNNs (also known as over-smoothing)~\cite{roth2024collapse} as well as %
previous conjectures on GNN vulnerability to bit flips being related to the injectivity of the MLP's modelling $f$~\cite{kummer2024attacking}. %
Furthermore, it has profound implications for GNNs such as GIN, which rely on moment-injectivity by modeling $f$ using $\MLP$'s, guaranteeing their ability to distinguish multisets representing 1-WL subtrees.

\label{ssec:vulgin}
We now investigate the relation between bit flips in the MLPs of neural moment-based GNNs and their expressivity on a certain, bounded graph dataset ${D}$. To this purpose, we first formulate a well-known result for easy reference later on. Let $\Phi^{(k)}$ be a GNN with $k$ layers, as in Equation~\eqref{eq:defginsimple}. Then, $\MLP^{(j)}_l, 1 \leq j \leq k$ denotes the $l$-layer $\MLP$ of $\Phi^{(k)}$'s $j$th layer. Let the $i$th layer of $\MLP^{(j)}_l$ be $\sigma \circ \mathbf{W}^{(j,i)} : {Q}^{n_{j,i}} \rightarrow {Q}^{m_{j,i}}$, $1 \leq i \leq l$. Clearly, if all layers $\sigma \circ \mathbf{W}^{(j,i)}$ of an $\MLP^{(j)}_l$ are injective, the $\MLP^{(j)}_l$ is injective and the following holds.
\lemma[Sufficient condition for the 1-WL expressivity of $\Phi^{(k)}$~\cite{leskovec2019powerful}]{
\label{lem:expressivegin}
If all layers \( \MLP^{(j)}_l \), for \( 1 \leq j \leq k \), of the GNN \( \Phi^{(k)} \) are injective and each layer's aggregation rule can distinguish a node's own features and the aggregated features from its neighbors, then \( \Phi^{(k)} \) is as expressive as the 1-WL test.
}\normalfont

Note that Lemma~\ref{lem:expressivegin} %
describes a sufficient condition. 
That is, a GNN might be maximally expressive on a certain, bounded dataset ${D}$ even if not all of it's $\MLP^{(j)}_l$'s are injective~\cite{kummer2024attacking}. 

Generalizing~\cite{puthawala2022globally}, we now define a general condition for the injectivity of $\sigma \circ \mathbf{W}^{(j,i)} : {Q}^{n_{j,i}} \rightarrow {Q}^{m_{j,i}}$ with linear transformation $\mathbf{W}^{(j,i)} \in {U}^{m_{j,i} \times n_{j,i}}$, $1 < n_{j,i} \leq m_{j,i}$, ${Q} \subset {U}$. Note the distinction between ${Q}$ and ${U}$, which we detail below. The domain ${Q}^{n_{j,0}}$ is defined as the set of all possible aggregates $\MLP^{(j)}_l$ might receive on a given set ${D}$ of graphs after $j$ layers. Thus, ${Q}^{n_{j,0}}$'s cardinality is bounded by the number of 1-WL colors occurring for any graph $G \in {D}$ after $j$ iterations $|{Q}^{n_{j,0}}| \leq |\{c^{(j)}_l(v) \mid v \in V(G), G \in {D} \}|$. %
Since any layer of $\MLP^{(j)}_l$ 
can produce at most as many distinguishable outputs as there are distinct inputs,
this also holds true for all $ 0 < i \leq l$. 
This implies that, likewise, the codomain ${Q}^{m_{j,i}}$'s cardinality is constrained to the cardinality of the set of the possible transformed aggregates. These constraints on the vector spaces ${Q}^{n_{j,i}}$ and ${Q}^{m_{j,i}}$ also bring limitations on the underlying field ${Q}$, which is implicitly constrained in its elements. These constraints do not apply to the vector space of $\mathbf{W}^{(j,i)} \in {U}^{m_{j,i} \times n_{j,i}}$, $1 < n_{j,i}$, which is why we define ${Q}$ as subset of ${U}$. 

Further, in the following, we assume $\sigma$ to be an analytic injective activation function unless stated otherwise. For such an activation function, a layer $\sigma \circ \mathbf{W}^{(j,i)}$ can be globally injective without the requirements on $\mathbf{W}^{(j,i)}$'s width and ${U}^{m_{j,i} \times n_{j,i}}$ applying to piece-wise-linear (PwL) activations~\cite{amir2024neural} such as ReLU~\cite{puthawala2022globally}.
We will discuss the special case of ReLU as an instance of PwL activations later in Section~\ref{sssec:sepcialrelu}.
\lemma[Equivalence of conditions for layer injectivity]{
\label{lem:mincond}
Let \( \mathbf{W}^{(j,i)} \in U^{m_{j,i} \times n_{j,i}} \) be a linear transformation with \( 1 < n_{j,i} \leq m_{j,i} \), and let \( R_{\mathbf{W}^{(j,i)}} = \{ \mathbf{w}_r \}_{r=0}^{m_{j,i}} \) denote the set of row vectors of \( \mathbf{W}^{(j,i)} \). Consider an injective elementwise activation function \( \sigma : U \rightarrow Q \). Then, the function \( \sigma \circ \mathbf{W}^{(j,i)} : Q^{n_{j,i}} \rightarrow Q^{m_{j,i}} \) is injective if and only if, for any distinct \( \mathbf{x}_u, \mathbf{x}_v \in Q^{n_{j,i}} \), there exists a row vector \( \mathbf{w}_r \in R_{\mathbf{W}^{(j,i)}} \) such that \( \langle \mathbf{x}_u, \mathbf{w}_r \rangle \neq \langle \mathbf{x}_v, \mathbf{w}_r \rangle \).
}\normalfont

If \( \mathbf{W}^{(j,i)} \) is of full rank, then \( \sigma \circ \mathbf{W}^{(j,i)} : Q^{n_{j,i}} \rightarrow Q^{m_{j,i}} \) satisfies the injectivity conditions outlined in Lemma~\ref{lem:mincond}.
The converse %
does not need to be true, though. Since the domain ${Q}^{n_{j,i}}$ is constrained %
by the number of possible aggregates at layer $(j,i)$, an injective mapping of ${Q}^{n_{j,i}}$ to ${Q}^{m_{j,i}}$ could also be facilitated if $\mathbf{W}^{(j,i)}$ does not have full rank.

We now provide the definition of a GNN that satisfies the outlined criteria for injectivity on a constrained domain and codomain. 
\definition[Maximally expressive GNN]{
\label{def:minimax}
A GNN \( \Phi^{(k)} \) is \emph{maximally expressive} if, for every layer of each \( \MLP^{(j)}_l \) in \( \Phi^{(k)} \), the function \( \sigma \circ \mathbf{W}^{(j,i)} \) satisfies the injectivity conditions outlined in Lemma~\ref{lem:mincond} and its aggregation rule can distinguish a node's own features and the aggregated features from its neighbors. Consequently, by Lemma~\ref{lem:expressivegin}, \( \Phi^{(k)} \) is as expressive as %
1-WL.%
}\normalfont

Although the subsequent analysis of node and graph-level expressivity (Sections~\ref{ssec:node_expressivity} and~\ref{ssec:graph_expressivity}) centers on %
neural moment-based GNNs such as GIN, it readily extends to other architectures. This includes Graph Convolutional Networks~\cite{welling2016semi} (GCN), by modeling GCN's %
transformations as 1-layer MLPs, which we include in our empirical evaluation, but (except for some of the special cases we discuss in Section~\ref{ssec:sepcial}) also translates to, e.g., Graph Attention Networks~\cite{velickovic2018graph} (GATs), which are likewise neural moment-based architectures, even if their sum is weighted by dynamically computed edge weights: their expressivity in the sense of their ability to distinguish non-isomorphic graphs is likewise bounded by the 1-WL algorithm and likewise intrinsically connected to the injectivity of the functions approximated by their trainable parameters (i.e. linear transformations combined with non-linear activations). %

\subsection{Node Level Expressivity}
\label{ssec:node_expressivity}
We explore the node-level expressivity of GNN $\Phi^{(k)}$, specifically lower and upper bounds on the number of bit flips in the $\MLP$s required to prevent a maximally expressive GNN from distinguishing all pairs of nodes in all graphs $G$ of a dataset ${D}$ that 1-WL distinguishes. The failure to distinguish even a single pair of WL-distinguishable nodes $u,v$ of any $G \in {D}$ %
suffices to break node-level 1-WL expressivity. In special cases, this could occur with a single bit flip: without constraints on node features or structures, inputs (aggregates) $\mathbf{x}_u, \mathbf{x}_v$ for nodes $u,v$ at layer $(j,i)$ may differ by only one element. If only one row $\mathbf{w}_r$ of $\mathbf{W}^{(j,i)}$ satisfies $\langle \mathbf{x}_u, \mathbf{w}_r \rangle \neq \langle \mathbf{x}_v, \mathbf{w}_r \rangle$, a single bit flip could map the distinguishing element to an identical value in the dot product, causing $\langle \mathbf{x}_u, \mathbf{w}_r \rangle = \langle \mathbf{x}_v, \mathbf{w}_r \rangle$ and further post-activation equality. By Lemmas~\ref{lem:expressivegin} and~\ref{lem:mincond}, maximal expressivity of $\Phi^{(k)}$ as in Definition~\ref{def:minimax} is no longer guaranteed.

Of greater interest is constructing a general upper bound. This can be achieved by selecting nodes $u,v$ at layer $(j,i)$ with the most distinctive aggregates $\mathbf{x}_u, \mathbf{x}_v$ among all nodes in the joint set $V_{\cup}({D}) = \bigcup_{G \in {D}} V(G)$. Note that the loss of distinguishability between $\mathbf{x}_u$ and $\mathbf{x}_v$ post-$\MLP^{j}_l$'s application implies a loss of $\MLP^{j}_l$'s injectivity, and thus, by %
Proposition~\ref{lem:mulinjlu}, the potential loss of layer $l$'s moment injectivity. In neural moment-based architectures, losing moment injectivity can prevent %
the distinction of multisets and thus %
node neighborhoods, which is required for node-level expressivity.
\theorem[Node-level upper bound]{
\label{thm:uppernode}
Let \( \Phi^{(k)}(\mathbf{A}, \mathbf{X}) \colon {D} \rightarrow {Q}^{m_k} \) be a maximally expressive GNN. The node-level expressivity of \( \Phi^{(k)} \) can be compromised by \( \mathcal{O}(d_{j,i} \cdot m_{j,i} \cdot b) \) bit flips, where \( d_{j,i} = \max_{\mathbf{x}_u, \mathbf{x}_v \in {Q}^{n_{j,i}}} \left \| \mathbf{x}_u - \mathbf{x}_v \right \|_0 \) for \( u,v \in V_{\cup}({D}) \) and \( b \) is the bit width.
}\normalfont

Theorem~\ref{thm:uppernode}, for which we provide computational complexity estimates in the supplementary material, effectively states that in the general case -- without constraints on graph structure or input features -- all neurons in a layer associated with differing aggregate elements must be zeroed out. While cases can be constructed where fewer bit flips suffice to negate the sufficient condition of $\mathbf{W}^{(j,i)}$ having full rank, this assumption does not hold in a worst-case analysis. Moreover, if $\max_{\mathbf{x}_u,\mathbf{x}_v \in {Q}^{n_{j,i}}}\left \| \mathbf{x}_u-\mathbf{x}_v \right \|_0 = n_{j,i}$, then $\mathcal{O}(d_{j,i} \cdot m_{j,i} \cdot b)$ degenerates to $\mathcal{O}(n_{j,i} \cdot m_{j,i} \cdot b)$, implying that distinguishability at layer $(j,i)$ is robust up to zeroing out every neuron in $\mathbf{W}^{(j,i)}$ for nodes $u$ and $v$. However, we can set limits on both node embedding diversity and variability, as shown in our analysis of graph-level expressivity and specific cases.

\subsection{Graph Level Expressivity}
\label{ssec:graph_expressivity} %
For graph-level expressivity, we explore bounds on the number of bit flips required to prevent the GNN from distinguishing two graphs that are distinguishable by 1-WL. Similar to node-level expressivity, we say $\Phi^{(k)}$ is maximally expressive for a dataset ${D}$ if it can distinguish any two graphs $G \not \simeq H$ that 1-WL distinguishes. Two graphs are distinguished by 1-WL if they have different colors or color multiplicities after $t$ iterations, meaning a GNN distinguishes them if they receive at least one differing node embedding or different embedding multiplicities. Thus, we examine the bit flips needed to align embeddings of non-isomorphic graphs. Since color (embedding) multiplicities trivially distinguish graphs of different order, we only consider graphs of equal order ($|V(G)| = |V(H)|$). Without assumptions on node features or graph structures, two graphs in ${D}$ could differ by a single node embedding at layer $(j,i)$, which could be degraded by a single bit flip (Section~\ref{ssec:node_expressivity}).%

To construct the graph-level upper bound (Theorem~\ref{thm:uppergraph}) from Theorem~\ref{thm:uppernode}, note that two graphs of equal order are indistinguishable by the GNN only if they share identical unique node embeddings with corresponding multiplicities. The number of unique node embeddings a GNN assigns at layer $(j,i)$ depends on both structure and node features. For MP-based GNNs, this can be bounded by the number of unique WL colors after $j$ iterations, upon which a difference measure is defined using the symmetric difference of multisets.
\definition[WL difference]{
\label{def:wldiff}
For two labeled graphs $G$ and $H$, we call $\Delta_{WL}(G,H,t) = |(C_l^{(t)}(G) \cup C_l^{(t)}(H)) \setminus (C_l^{(t)}(G) \cap C_l^{(t)}(H))|$ the \emph{WL difference} at iteration $t$.
}%
Then, the two equal order graphs $G_e, H_e \in {D}$ with the highest WL difference of any two graphs in ${D}$ 
at $k$ iterations are given by $G_e, H_e = \argmax_{G,H \in {D}, |V(G)| = |V(H)|}\Delta_{WL}(G,H,k)$.

\theorem[Graph-level upper bound]{
\label{thm:uppergraph}
Let \( \Phi^{(k)}(\mathbf{A}, \mathbf{X}) \colon {D} \rightarrow {Q}^{m_k} \) be a maximally expressive GNN. Consider the graphs \( G_e \) and \( H_e \) defined as \( G_e, H_e = \argmax_{G,H \in {D}, |V(G)| = |V(H)|} \Delta_{WL}(G,H,j) \), where \( e_j = \Delta_{WL}(G_e, H_e, j) \) represents the WL difference, and \( d_{j,i} = \max_{\mathbf{x}_u,\mathbf{x}_v \in {Q}^{n_{j,i}}} \left\| \mathbf{x}_u - \mathbf{x}_v \right\|_0 \) for \( u \in V(G_e) \) and \( v \in V(H_e) \). Then, the graph-level expressivity of \( \Phi^{(k)} \) can be degraded by \( \mathcal{O}(e_j \cdot d_{j,i} \cdot m_{j,i} \cdot b) \) bit flips, where \( b \) is the bit width.
}\normalfont

Similar to Theorem~\ref{thm:uppernode}, Theorem~\ref{thm:uppergraph} targets a sufficient condition for maximal expressivity defined in Definition~\ref{def:minimax}. 
For a GNN with a $\Readout$ function only using the embeddings of the last GNN layer, which would differ from Equation%
~\eqref{eq:defginreadout} but be closest to the WL algorithm, Theorem~\ref{thm:uppergraph} yields bounds for the guaranteed degradation of
expressivity for bit flips in the last layer. As for Theorem~\ref{thm:uppernode}, we provide computational complexity estimates for Theorem~\ref{thm:uppergraph} in the supplementary material.
\subsection{Special Cases}
\label{ssec:sepcial}
The preceding analysis becomes more specific when considering assumptions about activation functions, dataset, and vulnerable layers, refining our understanding of the system's robustness and expressivity
\paragraph{ReLU Activations}
\label{sssec:sepcialrelu}
While PwL-activated $\MLP$s can never induce a globally moment-injective function~\cite{amir2024neural}, they are highly popular. Specifically, $\ReLU$-activated $\MLP$s are subject to certain constraints on width and weights to ensure injectivity~\cite{puthawala2022globally}, even on constrained datasets, creating additional opportunities for an attacker to degrade GNN expressivity. Since width restrictions are more relevant in the global setting~\cite{puthawala2022globally}, which differs from our assumption of a bounded input domain, we focus on generally applicable weight limitations.
\lemma[Equivalence of conditions for layer injectivity (ReLU)]{
\label{lem:mincondrelu}
Let \( \mathbf{W}^{(j,i)} \in U^{m_{j,i} \times n_{j,i}} \) be a linear transformation with \( 1 < n_{j,i} \leq m_{j,i} \), and let \( R_{\mathbf{W}^{(j,i)}} = \{ \mathbf{w}_r \}_{r=0}^{m_{j,i}} \) denote the set of row vectors of \( \mathbf{W}^{(j,i)} \). Consider the ReLU activation function \( \ReLU : U \rightarrow Q^+ \). Then, the function \( \ReLU \circ \mathbf{W}^{(j,i)} : Q^{n_{j,i}} \rightarrow Q^{m_{j,i}} \) is injective if and only if for any distinct \( \mathbf{x}_u, \mathbf{x}_v \in Q^{n_{j,i}} \), there exists a row vector \( \mathbf{w}_r \in R_{\mathbf{W}^{(j,i)}} \) such that \( \langle \mathbf{x}_u, \mathbf{w}_r \rangle \neq \langle \mathbf{x}_v, \mathbf{w}_r \rangle \) and \( \langle \mathbf{x}_u, \mathbf{w}_r \rangle > 0 \) or \( \langle \mathbf{x}_v, \mathbf{w}_r \rangle > 0 \).
}\normalfont

The requirement for the dot product to be positive, as stated in Lemma~\ref{lem:mincondrelu}, makes the GNN highly vulnerable. Specifically, in a signed representation, flipping the sign bits in the target weights will suffice, letting ReLU zero out the activations. This reduces the node-level upper bound (Theorem~\ref{thm:uppernode}) from $\mathcal{O}(d_{j,i} \cdot m_{j,i} \cdot b)$ to $\mathcal{O}(d_{j,i} \cdot m_{j,i})$, independent of bit width $b$, as there is at most one sign bit per weight element. Similarly, it reduces the graph-level upper bound (Theorem~\ref{thm:uppergraph}) from $\mathcal{O}(e_j \cdot d_{j,i} \cdot m_{j,i} \cdot b)$ to $\mathcal{O}(e_j \cdot d_{j,i} \cdot m_{j,i})$.
\paragraph{Attacks on the First Layer} 
A common assumption in %
literature is that node features are one-hot encodings of graph node labels~\cite{leskovec2019powerful}, as seen in many real-world benchmark datasets~\cite{morris2020tudataset, hu2020open}. We examine how this encoding influences GNN expressivity vulnerability, particularly in the first layer. Assume the dataset has \( g \) unique node labels, so the dimension of the one-hot encoded vectors is \( g \). Let \( d = \max_{G \in {D}}(\deg_{\max}(G)) \) denote the maximum node degree across all graphs in the dataset. The impact of these characteristics on the first layer of a GNN is captured in the following corollary.
\corollary[Node-Level upper bound (first layer)]{
\label{lem:first_layer_vulnerability}
Let \(\Phi^{(k)}\) be a GNN and let the node features of all $G \in {D}$ be represented as one-hot encodings of the graph's node labels. Assume the dataset has \( g \) unique node labels, and let \( d = \max_{G \in {D}}(\deg_{\max}(G)) \) denote the maximum node degree across all graphs in the dataset. Then, the expressivity of the first layer \(\MLP^{(1)}_l\) can be compromised by flipping \( \mathcal{O}(m_{1,1} \cdot b \cdot nz) \) bits, where \( nz = \min(2 \cdot d, n_{1,1}) \), \( m_{1,1} \) is the number of output dimensions, and \( b \) is the bit width.
}\normalfont

The required number of bit flips decreases as the difference between \( d \) and \( g \) increases, since \( n_{1,1} = g \) is fixed for the first layer. This reflects a specialized case of the general node-level upper bound in Theorem~\ref{thm:uppernode}, where vulnerability is heightened by the sparse, structured nature of one-hot encoded inputs. For the generalization from Theorem~\ref{thm:uppernode} to Theorem~\ref{thm:uppergraph}, note that Theorem~\ref{thm:uppergraph} (like Theorem~\ref{thm:uppernode}, without assumptions on node features or attacked layer) derives the number of elements to zero out per neuron as $d_{j,i} = \max_{\mathbf{x}_u,\mathbf{x}_v \in {Q}^{n_{j,i}}}\left | \mathbf{x}_u-\mathbf{x}_v \right |_0$, $u \in V(G_e), v \in V(H_e)$ with $G_e, H_e= \text{argmax}_{G,H \in {D}, |V(G)| = |V(H)|}\Delta_{WL}(G,H,j)$. Assuming $\mathbf{W}^{(1,1)}$ is attacked and node features are one-hot encodings, we can bound $\left | \mathbf{x}_u-\mathbf{x}_v \right |_0$ by the same argument.

\paragraph{The Impact of Homophily}
To refine the node-level upper bound in Theorem~\ref{thm:uppernode}, particularly for attacks on the input layer with one-hot encoded inputs, we incorporate the dataset's graph homophily ratio. The homophily ratio quantifies the likelihood that connected nodes share the same label (not to be confused with the class labels of nodes), measuring similarity within the graph structure. This relationship is formalized in the following corollary.
\corollary[Node-Level upper bound (homophily)]{
\label{lem:homophily_vulnerability}
Let \( H_G \) denote the homophily ratio of a graph \( G(V,E) \) with node labels \( l \), defined as \( H_G = |E(G)|^{-1}|\{(v_i, v_j) \in E(G) \mid l(v_i) = l(v_j)\}| \). For a dataset \( D \), the average homophily ratio is \( H_{{D}} = |{D}|^{-1}\sum_{G \in {D}} H_G \). Let \( P_{{D}} \) denote the probability that two randomly chosen nodes from \( V_{\cup}(D) \) are connected, given by \( P_{{D}} = \sum_{G \in {D}} |E(G)|\cdot\binom{|V(G)|}{2}^{-1} \). The number of bit flips required to make two nodes \( u, v \in V_{\cup}(D) \) indistinguishable in the first layer is bounded by \( \mathcal{O}(m_{1,1} \cdot b \cdot nz_H) \), where \( nz_H = \min(2 \cdot d \cdot (1-H_{{D}}) \cdot (1-P_{{D}}), n_{1,1}) \), \( d \) is the maximum node degree, \( m_{1,1} \) is the output dimension of the first layer, and \( b \) is the bit width.
}\normalfont

\begin{figure*}[t] %
    \begin{center}
    \centerline{\includegraphics[width=1.0\textwidth, trim={0.025cm 0.32cm 0.025cm 0.17cm}, clip]{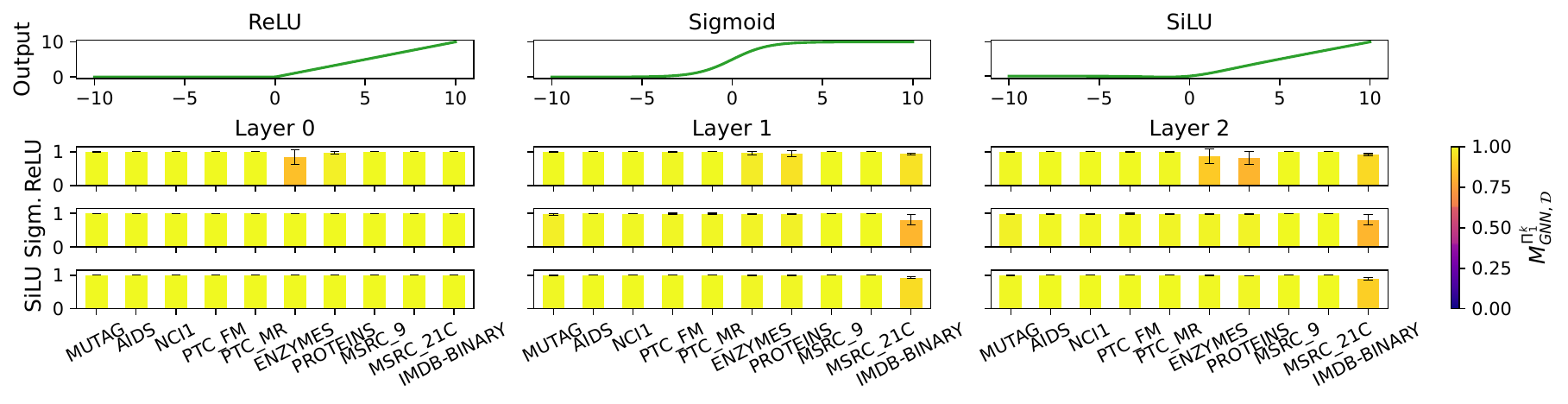}} 
    \caption{\label{fig:barplot_umr} Visualization of $\mu \pm \sigma$ of the metric $M^{\Pi_1^k}_{GNN, {D}}$ (bottom) computed from the 20250 unperturbed (clean) models used in the attack runs for ReLU, Sigmoid, SiLU (top), aggregated across 3-layer GIN/GCN and DS.}
    \end{center}
    \vskip -0.2in
\end{figure*}
The homophily ratio \( H_D \) influences the sparsity of the aggregates: a lower homophily ratio increases the likelihood that neighboring nodes have different labels, leading to fewer common non-zero entries in the embeddings of connected nodes. Since \( nz_H \) is typically less than the general case \( nz \) (unless \( H_D = P_D = 1 \), which is rare), the refined bound \( \mathcal{O}(m_{1,1} \cdot b \cdot nz_H) \) usually offers a tighter estimate for vulnerability when homophily is significant. This analysis shows how the dataset's structure and homophily can exacerbate GNN expressivity vulnerability, particularly with one-hot encoded inputs, refining the bounds in Theorem~\ref{thm:uppernode} and allowing extension to Theorem~\ref{thm:uppergraph}.

Note that the preceding formal analysis of the impact of homophily and attacks on the first layer might not extend to GAT, as the pre-transformation application of attention parameters and nonlinear transformations during aggregation might reduce or entirely negate these influence factors, and more analysis is needed to fully clarify their impact on these architectures.
\section{EXPERIMENTS}
\begin{figure*}%
    \begin{center}
    \centerline{\includegraphics[width=1.0\textwidth, trim={0.025cm 0.075cm 0.025cm 0.17cm}, clip]{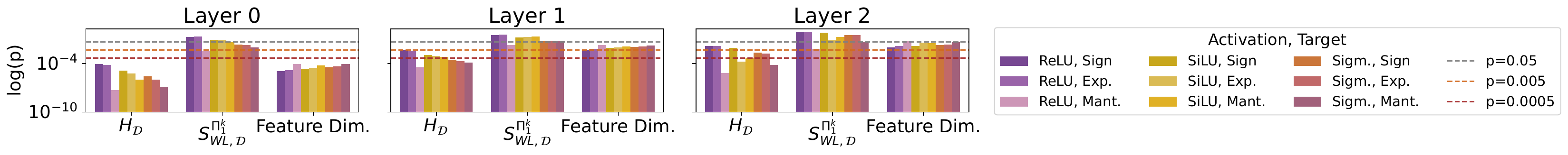}}
    \caption{\label{fig:pvals_corr} Significance levels of Spearman correlation of dataset properties $S_{WL, {D}}^{(k)}$,  $H_{{D}}$ and feature dimensionality
    with $\Delta Exp$ computed across all runs and datasets for exponent, sign and mantissa target bits in Sigmoid, ReLU and SiLU activated 3-layer GIN/GCNs and DS (101250 runs in total).}
    \end{center}
    \vskip -0.2in
\end{figure*}
\label{sec:experiments}
We design our experiments to investigate key research questions (RQs) derived from our theoretical analysis, using 10 real-world %
datasets from TUDataset~\cite{morris2020tudataset}. These datasets, detailed in the supplementary material, are popular in contemporary research, feature one-hot or binary encoded node labels, and cover a range of common tasks in drug development~\cite{xiong2021novo, rossi2020proximity}, bioinformatics~\cite{borgwardt2005protein}, object recognition~\cite{rossi2015repo}, and social networks.%

Specifically, we explore 

(\textbf{RQ1}) whether the vulnerable bits in the floating-point representation depend on the activation function,

(\textbf{RQ2}) whether a loss of expressivity is linked to a loss of injectivity in the GNNs' MLPs, and

(\textbf{RQ3}) whether GNN resilience is connected to the graph partitioning induced by the WL coloring. 

Moreover, we examine

(\textbf{RQ4}) whether resilience to first layer attacks relates to homophily and feature dimensionality.

We assess our RQs on GIN (2 hidden layers per $\MLP$) due to its direct relation to our theory as well as on an architecture directly derived from DeepSets implementing Equation~\ref{eq:defginsimple}, which we abbreviate as DS (likewise 2 hidden layers per $\MLP$). To demonstrate that (as claimed in Section~\ref{ssec:vulgin}) our theory extends to GCN, representing the class of spectral GNNs, we include GCN in our assessment. 
We use ReLU, Sigmoid and SiLU~\cite{apicella2021survey} activations %
and initialize network parameters $\mathbf{W}^{(j,i)}$ %
randomly from a uniform distribution $\mathcal{U}(-\sqrt{\frac{1}{m_{j_i}}},\sqrt{\frac{1}{m_{j_i}}})$ following common variance scaling initialization schemes~\cite{glorot2010understanding, he2015delving}. 
As GNNs are maximally expressive for randomly initialized weights in almost all cases~\cite{amir2024neural} and our theoretical upper bounds, including those taking influence factors such as homophily into account, hold for all possible weights, we do not train our GNNs. Bits are flipped semi-randomly in either the sign, exponent, or the mantissa of the \texttt{FLOAT32} represented weights, to which we refer as target bits: specifically and in line with our theory (Section~\ref{ssec:vulgin}), we randomly induce $0 \rightarrow 1$ flips in sign bits and $1 \rightarrow 0$ flips in mantissa and exponent bits. For each experiment, we randomly initialize 5 GNN instances and repeat the induction of a certain percentage of bit flips in the target bits 5 times per instance. This results in 25 runs per experiment in total, reducing stochastic effects. Our experiments\footnote{Our code is available in our repository at
\url{https://github.com/lorenz0890/exproaistats}.
} took 8 weeks with three parallel workers to conclude and were conducted on a local server equipped with an NVIDIA H100 PCIe GPU (80GB VRAM), an Intel Xeon Gold 6326 CPU (500GB RAM) and a 1TB SSD.%

\paragraph{Metrics}
We measure graph level expressivity (abbreviated $Exp$) as the percentage of non-isomorphic graphs of a dataset for which the GNN's final MP layer outputs node embeddings such that their sum is distinguishable 
for \texttt{FLOAT32} ($\epsilon_{mach} = 1.19\times 10^{-7}$), which necessitates that they have at least one distinguishable node embedding element. We do not introduce an additional tolerance level, as the impact of numerical error propagation on GNN expressivity has not yet been sufficiently investigated.
This is equal to the assumption of a $\Readout$ function that only encodes the last layer's embeddings and enables us to pinpoint how bit flips in specific layers affect expressivity.
We further measure the degree of refinement of the partitions induced by the GNN embeddings, as well as the 1-WL colors. Specifically, we use the \emph{average WL subdivision ratio} $S_{WL, {D}}^{(t)} = |{D}|^{-1} \sum_{G \in {D}}\frac{|C_l^{(t)}(G)|}{|C_l^{(t-1)}(G)|}$, $t \geq 1$, and the \emph{average GNN subdivision ratio} $S_{GNN, {D}}^{(k)} = |{D}|^{-1} \sum_{(\mathbf{A}, \mathbf{X}) \in {D}}\frac{|\Phi^{(k)}(\mathbf{A}, \mathbf{X})|_{\delta}}{|\Phi^{(k-1)}(\mathbf{A}, \mathbf{X})|_{\delta}}$, $k \geq 1$, of a given dataset ${D}$ to measure 1-WL’s and the GNN's embeddings partition diversity, respectively, whereby the $|\cdot|_{\delta}$ describes the number of elementwise distinguishable node embeddings vectors of a given graph up to ${\epsilon_{mach}}$.
\begin{figure*}%
    \begin{center}
    \centerline{\includegraphics[width=1.0\textwidth, trim={0.025cm 0.75cm 0.025cm 0.17cm}, clip]{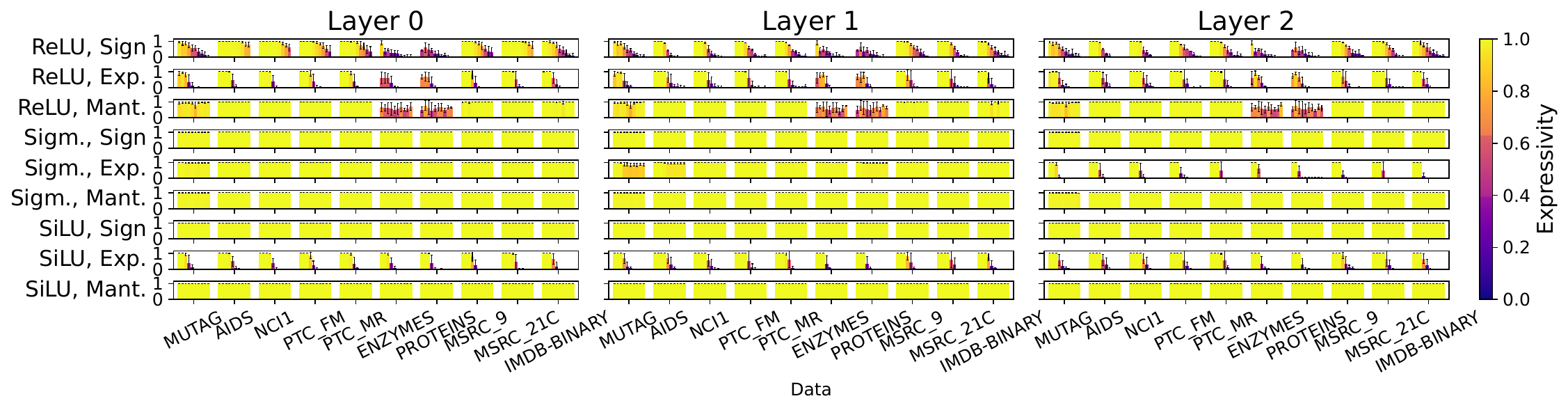}}
    \caption{\label{fig:res} Flips in exponent, sign and mantissa bits of Sigmoid, ReLU and SiLU activated 3-layer GIN/GCNs and DS. Each group of $10$ bars represents %
    expressivity (i.e. the fraction of distinguishable non-isomorphic graphs of the given dataset) after (left to right) %
    1\% to 95\% bits flipped in a certain component of the \texttt{FLOAT32} representation of the weights in a layer. Each bar represents $\mu \pm \sigma$ computed from 25 runs (101250 runs in total).} %
    \end{center}
    \vskip -0.2in
\end{figure*} 

For specific $k$, it is possible that $S_{GNN, {D}}^{(k)} \geq S_{WL, {D}}^{(k)}$ since the GNN could recover from previous failures to distinguish certain unfolding trees if a layer's %
preceding functions were not injective~\cite{kummer2024attacking}. However, it generally holds that $\prod_{k=1}^{n}S_{GNN, {D}}^{(k)} = S_{GNN, {D}}^{\Pi^n_1} \leq S_{WL, {D}}^{\Pi^n_1} = \prod_{k=1}^{n}S_{WL, {D}}^{(k)}$ since WL is the upper bound of MP GNN expressivity.%

Since the injectivity of a function by itself %
can not be quantified, we introduce a substitute %
to determine the degree to which the $\MLP^{(j)}_l$ of a GNN are capable to produce a 1-to-1 mapping of their input domain to their output domain. That is, to measure the mapping of functions learned by $\MLP^{(j)}_l$'s at the $j$th layer, we measure the ratio of uniquely mapped inputs across the entire dataset ${D}$ using the \emph{unique mapping ratio} $M^{(j)} = {|{H}^{(j)}|_{\delta}}|{Z}^{(j)}|_{\delta}^{-1}$, which represents the ratio of inputs uniquely mapped  (up to ${\epsilon_{mach}}$) by the $\MLP^{(j)}_l$, whereby ${Z}^{(j)}$ is the set of all inputs the $\MLP^{(j)}_l$ receives on a dataset, representing node embeddings of graphs in ${D}$ at layer $j-1$, and ${H}^{(j)}$ the set of all outputs $\MLP^{(j)}_l$ produces for these inputs.

Clearly, $\MLP^{(j)}_l$ is injective w.r.t.~to its input domain if $M^{(j)} = 1$. We compute an aggregate score as $M^{\Pi_1^k}_{GNN, {D}}= \prod_{j=1}^k M^{(j)}$. Then, $M^{\Pi_1^k}_{GNN, {D}} = 1$ implies injectivity of all layers, hence Lemma~\ref{lem:expressivegin}'s criteria are satisfied and consequently, GNN can be maximally expressive in the sense of Definition~\ref{def:minimax}. %
We find $M^{\Pi_1^k}_{GNN, {D}}$ is nearly $1$ for randomly initialized GNNs in most cases, see Figure~\ref{fig:barplot_umr}, which is in line with the results of \citet{amir2024neural}. %
Note that $M^{\Pi_1^k}_{GNN, {D}} = 1$ implies $S_{GNN}^{\Pi^k_1}({D}) = S_{WL}^{\Pi^k_1}({D})$ for maximally expressive MP based GNNs such as GIN (but not vice versa).

\subsection{Results}
\label{ssec:results}
Our empirical results (Figure~\ref{fig:res}) confirm our theoretical predictions.

(\textbf{RQ1}) ReLU-activated GNNs are particularly susceptible to bit flips, including sign bit flips, distinguishing them from Sigmoid- or SiLU-activated GNNs, as predicted by Lemmas~\ref{lem:mincond} and~\ref{lem:mincondrelu}. This vulnerability extends to upper layers, where expressivity degradation cannot be guaranteed for injective activations (Section~\ref{ssec:sepcial}), even for $M^{\Pi_1^k}_{GNN, {D}} < 1$. Attacks on the last layer's exponents degrade expressivity across all GNNs, datasets, and activations. Upper-layer vulnerability to bit flips in the exponent was only observed for non-injective activations ReLU and SiLU. 

(\textbf{RQ2}) A relative decrease in injectivity is linked with expressivity degradation: with a Spearman correlation coefficient of $0.5209$ across all experiments, there is a strong correlation between $\Delta M^{\Pi^k_1}_{GNN, {D}}$ and $\Delta Exp$, significant at $p = 1.2 \cdot 10^{-49}$ ($p < 0.005$). This aligns with the predicted potential loss of moment injectivity (Proposition~\ref{lem:mulinjlu}).
While Spearman correlation is robust to non-linear relationships, Pearson correlation showed similar significance only for first-layer bit flips, with no significant correlations in deeper layers, indicating a more complex relationship.

(\textbf{RQ3}) %
No general direct link between resilience and the granularity of the WL color partitioning (represented by $S^{\Pi^k_1}_{WL, {D}}$) of a dataset's graphs is established for all layers and activations but only specific combinations thereof (Figure~\ref{fig:pvals_corr}). 
However, the reduction in GNN subdivision ratio $\Delta S^{\Pi^k_1}_{GNN, {D}}$ is highly correlated with $\Delta Exp$, with a coefficient of $0.7390$ at $p = 1.9 \cdot 10^{-319}$, indicating an indirect link to $S^{\Pi^k_1}_{GNN, {D}}$'s upper bound $S^{\Pi^k_1}_{WL, {D}}$ and suggesting that practical resilience is influenced by additional factors. %

(\textbf{RQ4}) $H_{{D}}$ and feature dimensionality show a low but highly significant ($p < 0.005$) correlation with $\Delta Exp$ for first layer bit flips (Figure~\ref{fig:pvals_corr}) as expected by Corollaries~\ref{lem:first_layer_vulnerability} and~\ref{lem:homophily_vulnerability}, with %
coefficients $0.099$ and $-0.088$, respectively. This suggests that higher homophily correlates with increased expressivity loss, while increased feature dimensionality is protective. No significant correlation between $H_{{D}}$ or feature dimensionality and $\Delta Exp$ is found in deeper layers.

\paragraph{Implications for Practitioners}
Our experiments confirm the importance of homophily, feature dimensionality, and the GNN/1-WL subdivision ratio in influencing GNN resilience, indirectly validating the theoretical framework and suggesting that the broad upper bounds provided by our main Theorems~\ref{thm:uppernode} and~\ref{thm:uppergraph} capture critical aspects relevant to real-world datasets. 
While expressivity loss may not always reduce practical performance, whenever bit-flip-induced expressivity loss does degrade predictive quality, our bounds on the required number of bit flips remain valid. In particular, any performance drop due to expressivity loss occurs with fewer bit flips than our worst-case estimates.

The observed correlations underscore the significance of these factors in designing robust GNNs, especially where expressivity is crucial. As our analysis applies broadly to any architecture relying on injective neural moments to distinguish multisets, our results
offer actionable insights for practitioners: ReLU-activated GNNs' expressivity is more vulnerable to bit flips, particularly due to feature dimensionality and encoding schemes. To enhance robustness, practitioners could adopt SiLU activations as a resilient alternative to ReLU or increase feature dimensionality through pre-coloring techniques, such as~\cite{gonzalez2022beyond, cai2018simple}. %
Additionally, transforming one-hot encoded features into a dense format while preserving linear independence could maintain expressivity and improve robustness (see Section~\ref{ssec:neumom}).

\section{CONCLUSION}
We establish theoretical foundations to understand the resilience of neural moment-based GNNs against BFAs and provide provable bounds for the number of bit flips required to degrade expressivity.
We show that GNN robustness to BFAs is fundamentally influenced by architectural properties, the graph dataset's structural properties, and the activation functions employed. We find that ReLU-activated GNNs are particularly susceptible to BFAs, whereas employing alternative activation functions, such as SiLU or Sigmoid, enhances resilience. We further find that GNN resilience is significantly influenced by the dataset's homophily and feature dimensionality. Specifically, higher homophily tends to exacerbate the impact of bit flips, whereas higher feature dimensionality increases resilience. These insights underline the importance of considering both graph topology and feature space design when developing GNNs for applications where bit flip-robustness is paramount. 
Future work will explore the relationship between these theoretical factors and practical considerations, such as class label distributions, model training, optimization algorithms, learning rates, weight initialization, and regularization, deepening our understanding of practical vulnerabilities and their mitigation.

\section{Acknowledgments}
This work was supported by the Vienna Science and Technology
Fund (WWTF) [10.47379/VRG19009].%

\bibliographystyle{apalike}
\bibliography{aistats}

\newpage
\newpage
\appendix
\section{PROOFS}
\label{apx:proofs}
\paragraph{\LK{Proof of Proposition~\ref{lem:mulinjlu}}} 
We prove by contradiction that $(\forall A \in \Omega_n^d |  \sum_{\mathbf{x} \in S(A)} f(\mathbf{x}) \lambda_{\mathbf{x}} = \mathbf{0} \Rightarrow \forall \mathbf{x} \in A | \lambda_{\mathbf{x}} = 0, \lambda_{\mathbf{x}} \in V) \Longrightarrow (\forall A, B \in \Omega_n^d | \Hat{f}(A) = \Hat{f}(B) \Rightarrow A = B$, $n \geq 2)  $ 

Assume the negation of the necessary condition: 
\\ (1.) $ \exists A, B \in \Omega_n^d | A \neq B \wedge \Hat{f}(A) = \Hat{f}(B) $. 
\\\\ Then, through transformations, we reach: 
\\(2.) $\Hat{f}(A) = \Hat{f}(B) \Leftrightarrow$ 
\\(3.) $\sum_{\mathbf{x} \in A} f(\mathbf{x}) = \sum_{\mathbf{x} \in B} f(\mathbf{x}) \Leftrightarrow$ 
\\(4.) $\sum_{\mathbf{x} \in S(A)} m_A(\mathbf{x})f(\mathbf{x}) - \sum_{\mathbf{x} \in S(B)} m_B(\mathbf{x})f(\mathbf{x}) = \mathbf{0} \Leftrightarrow$ 
\\(5.) $\sum_{\mathbf{x} \in S_{\cup}} m_A(\mathbf{x})f(\mathbf{x}) - \sum_{\mathbf{x} \in S_{\cup}} m_B(\mathbf{x})f(\mathbf{x}) = \mathbf{0} \Leftrightarrow$
\\(6.) $\sum_{\mathbf{x} \in S_{\cup}}(m_A(\mathbf{x}) - m_B(\mathbf{x}))f(\mathbf{x}) = \mathbf{0}$ 
\\ with  $S_{\cup} = S(A \cup B)$ but without guarantee for $S_{\cup} \in \Omega_n^d$ since $|S_{\cup}| \leq 2n$.
\\\\ However, because $A \neq B \Rightarrow |A - B| > 0$, it follows that
\\ (7.) $\exists \mathbf{x} \in S_{\cup} | (m_A(\mathbf{x}) - m_B(\mathbf{x})) \neq 0$ and consequentially, to satisfy (6.), that 
\\ (8.) $\exists \mathbf{y} \in S_{\cup}, \mathbf{y} \neq \mathbf{x}, (m_A(\mathbf{y}) - m_B(\mathbf{y})) \neq 0 | (m_A(\mathbf{x}) - m_B(\mathbf{x}))f(\mathbf{x}) + (m_A(\mathbf{y}) - m_B(\mathbf{y}))f(\mathbf{y}) = \mathbf{0}$.
\\\\
Note we assume $f(\mathbf{y}) \neq f(\mathbf{x})$ in (8.) because $f(\mathbf{y}) = f(\mathbf{x})$ for $\mathbf{x} \neq \mathbf{y}$, would %
directly contradict $\forall A \in \Omega_n^d |  \sum_{\mathbf{x} \in S(A)} f(\mathbf{x}) \lambda_{\mathbf{x}} = \mathbf{0} \Rightarrow \forall \mathbf{x} \in A | \lambda_{\mathbf{x}} = 0, \lambda_{\mathbf{x}} \in V$. %
\\
\\ To proceed, choosing $M = \multiset{\mathbf{y}, \mathbf{x}}{} \subseteq S_{\cup}$, $|M| = 2 \leq n$ and thus $M \in \Omega_n^d$, leads to
\\ (9.) $f(\mathbf{x})\lambda_{\mathbf{x}} + f(\mathbf{y}) \lambda_{\mathbf{y}} = \mathbf{0}$ with $\lambda_{\mathbf{x}} = (m_A(\mathbf{x}) - m_B(\mathbf{x})) \neq 0$ and $\lambda_{\mathbf{y}} = - (m_A(\mathbf{y}) - m_B(\mathbf{y})) \neq 0$, $\mathbf{x},\mathbf{y} \in M$, contradicting the sufficient condition that $\forall A \in \Omega_n^d |  \sum_{\mathbf{x} \in S(A)} f(\mathbf{x}) \lambda_{\mathbf{x}} = \mathbf{0} \Rightarrow \forall \mathbf{x} \in A | \lambda_{\mathbf{x}} = 0$.

\qed
\paragraph{Proof of Lemma~\ref{lem:mincond}}
If for arbitrary $\mathbf{x}_u, \mathbf{x}_v \in {Q}^{n_{j,i}}, \mathbf{x}_u \neq \mathbf{x}_v$ it holds that $\exists \mathbf{w}_r \in {R}_{\mathbf{W}^{(j,i)}}$ s.t. $\langle \mathbf{x}_u, \mathbf{w}_r \rangle \neq \langle \mathbf{x}_v, \mathbf{w}_r \rangle $, then $\sigma(\mathbf{W}^{(j,i)}\mathbf{x}_u) \neq \sigma(\mathbf{W}^{(j,i)}\mathbf{x}_v)$ follows directly from $\sigma$'s injectivity. Conversely, assume $ \forall \mathbf{x}_u, \mathbf{x}_v \in {Q}^{n_{j,i}}$ with $\mathbf{x}_u \neq \mathbf{x}_v$ it holds that $\sigma(\mathbf{W}^{(j,i)}\mathbf{x}_u) \neq \sigma(\mathbf{W}^{(j,i)}\mathbf{x}_v)$, then $\exists \mathbf{w}_r \in {R}_{\mathbf{W}^{(j,i)}}$ s.t. $\sigma(\langle \mathbf{x}_u, \mathbf{w}_r \rangle) \neq \sigma(\langle \mathbf{x}_v, \mathbf{w}_r \rangle)$. Because of $\sigma$'s injectivity, $\neg(\sigma(\langle \mathbf{x}_u, \mathbf{w}_r \rangle)\neq \sigma(\langle \mathbf{x}_v, \mathbf{w}_r \rangle))$ would imply $\langle \mathbf{x}_u, \mathbf{w}_r \rangle = \langle \mathbf{x}_v, \mathbf{w}_r \rangle $ and further $\mathbf{x}_u = \mathbf{x}_v$. This would stand in contradiction to the assumption $\mathbf{x}_u \neq \mathbf{x}_v$, such that $ \exists \mathbf{w}_r \in {R}_{\mathbf{W}^{(j,i)}}$ for which $\langle \mathbf{x}_u, \mathbf{w}_r \rangle \neq \langle \mathbf{x}_v, \mathbf{w}_r \rangle $ follows. 
\qed

\paragraph{Proof of Theorem~\ref{thm:uppernode}}
Let $\Phi^{(k)}$ be a maximally expressive GNN (Definition~\ref{def:minimax}) with $k$ layers as in~\eqref{eq:defginsimple}. That is, each layer $\sigma \circ \mathbf{W}^{(j,i)}$ of each $\MLP^{(j)}_l$ of $\Phi^{(k)}$ satisfies the conditions of Lemma~\ref{lem:mincond}. Let $(j,i)$ be an arbitrary layer. Let $\mathbf{x}_u, \mathbf{x}_v \in {Q}^{n_{j,i}}, \mathbf{x}_u \neq \mathbf{x}_v$ be chosen such that the number of indices where $\mathbf{x}_u, \mathbf{x}_v$ differ elementwise $d_{j ,i} = \max_{\mathbf{x}_u,\mathbf{x}_v \in {Q}^{n_{j,i}}}\left \| \mathbf{x}_u-\mathbf{x}_v \right \|_0$ is maximal for $u,v \in V_{\cup}({D})$.
Then, at most $d_{j ,i}$ weight elements of $\mathbf{W}^{(j,i)}$ with $b$ bits each would have to be modified. In the worst case, that is setting the $d_{j ,i}$ weights elements associated with the $d$ indices where $\mathbf{x}_u, \mathbf{x}_v$ differ to $0$ at the cost of at most $b$ bit flips in each of the $m_{j,i}$ rows $\mathbf{w}_r$, amounting to at most $d_{j ,i} \cdot m_{j,i} \cdot b$ bit flips after which $\Phi^{(k)}$'s maximal node-level expressivity can not be guaranteed anymore. 
\qed
\paragraph{Proof of Theorem~\ref{thm:uppergraph}}
Let $G_e, H_e = \argmax_{G,H \in {D}}\Delta_{WL}(G,H,j)$ and $e_j = \Delta_{WL}(G_e, H_e, j)$.
That is, $G_e$ and $H_e$ are chosen from ${D}$ such that they are structurally most differentiable by WL after $j$ iterations, $e_j > 0 \Rightarrow G_e \not \simeq H_e$.
Thus, also the outputs of the $\MLP^{(j)}_l$, $j \leq k$, of a GNN $\Phi^{(k)}$ satisfying Definition~\ref{def:minimax} will differ in at most $e_j$ embedding vectors. Then, as follows from Theorem~\ref{thm:uppernode}, a given layer $(j,i)$ will require at most $d_{j, i} \cdot m_{j,i} \cdot b$ bit flips with $d_{j,i} = \max_{\mathbf{x}_u,\mathbf{x}_v \in {Q}^{n_{j,i}}}\left \| \mathbf{x}_u-\mathbf{x}_v \right \|_0$, $u \in V(G_e), v \in V(H_e)$, to make a single pair of the $e_j$ WL distinguishable nodes of $G_e$ and $H_e$ receive the same embeddings. If we repeat this step until it is guaranteed that none of the nodes of $G_e$ and $H_e$ can be distinguished pairwise at layer $(j, i)$ of $\Phi^{(k)}$, we require at most $e_j \cdot d_{j, i} \cdot m_{j,i} \cdot b$ bit flips. 
\qed

\paragraph{Proof of Lemma~\ref{lem:mincondrelu}}
If for arbitrary $\mathbf{x}_u, \mathbf{x}_v \in {Q}^{n_{j,i}}, \mathbf{x}_u \neq \mathbf{x}_v$ it holds that $\exists \mathbf{w}_r \in {R}_{\mathbf{W}^{(j,i)}}$ s.t. $\langle \mathbf{x}_u, \mathbf{w}_r \rangle \neq \langle \mathbf{x}_v, \mathbf{w}_r \rangle $ and $\langle \mathbf{x}_u, \mathbf{w}_r \rangle \geq 0, \langle \mathbf{x}_v, \mathbf{w}_r\rangle \geq 0$, then $\ReLU(\mathbf{W}^{(j,i)}\mathbf{x}_u) \neq \ReLU(\mathbf{W}^{(j,i)}\mathbf{x}_v)$ follows directly from ReLUs acting as identity function on positive inputs. Conversely, assume $ \forall \mathbf{x}_u, \mathbf{x}_v \in {Q}^{n_{j,i}}$ with $\mathbf{x}_u \neq \mathbf{x}_v$ it holds that $\ReLU(\mathbf{W}^{(j,i)}\mathbf{x}_u) \neq \ReLU(\mathbf{W}^{(j,i)}\mathbf{x}_v)$, then $\exists \mathbf{w}_r \in {R}_{\mathbf{W}^{(j,i)}}$ s.t. $\ReLU(\langle \mathbf{x}_u, \mathbf{w}_r \rangle) \neq \ReLU(\langle \mathbf{x}_v, \mathbf{w}_r \rangle)$. Because ReLU acts as identity for positive inputs and zeros out negative inputs, it follows that $\langle \mathbf{x}_u, \mathbf{w}_r \rangle > 0 \vee  \langle \mathbf{x}_v, \mathbf{w}_r\rangle > 0$ and $\langle \mathbf{x}_u, \mathbf{w}_r \rangle \neq \langle \mathbf{x}_v, \mathbf{w}_r \rangle$. 
\qed

\paragraph{Proof of Corollary~\ref{lem:first_layer_vulnerability}}
Assume that the dimension of the one-hot encoded node label vectors is \( g \). After sum aggregation in the first layer's \(\MLP^{(1)}_l\), the number of non-zero elements (i.e., the number of distinct contributions from neighboring nodes) for any given node is at most \( \min(d, n_{1,1}) \), where \( n_{1,1} \) is the input dimension to the first layer. Consequently, the number of zero elements in the aggregation is at least \( \max(g - d, 0) \).

Consider two nodes \( u, v \in V_{\cup}({D}) \) that have the largest number of common non-zero embedding entries, denoted as \( nz = \min(2 \cdot d, n_{1,1}) \). To make these nodes indistinguishable after the application of \(\MLP^{(1)}_l\), it follows from Theorem~\ref{thm:uppernode} that \( \mathcal{O}(m_{1,1} \cdot b \cdot nz) \) bit flips would be required in the first layer's \(\MLP \). 
\qed
\paragraph{Proof of Corollary~\ref{lem:homophily_vulnerability}}
For a graph \( G(V,E) \) with node labels \( l \), the homophily ratio \( H_G \) quantifies the probability that two connected nodes share the same label. This ratio is averaged over a dataset \( D \) to yield the average homophily ratio \( H_D \), which indicates the expected probability that any two connected nodes in \( V_{\cup}(D) \) share the same label.

Next, we consider the probability \( P_D \), representing the likelihood that two randomly chosen nodes from \( V_{\cup}(D) \) are connected. For one-hot encoded input vectors, both the connectivity and homophily influence the non-zero entries in the node embeddings. Specifically, for any pair of nodes \( u, v \in V_{\cup}(D) \), the number of non-zero entries in their embeddings can be estimated as \( nz_H = \min(2 \cdot d \cdot (1-H_{{D}}) \cdot (1-P_{{D}}), n_{1,1}) \), where \( d \) is the maximum node degree, and \( n_{1,1} \) is the input dimension to the first layer. Consequently, the number of bit flips required to make these nodes indistinguishable in the first layer decreases. Thus, this leads to a more vulnerable GNN expressivity, specifically bounded by \( \mathcal{O}(m_{1,1} \cdot b \cdot nz_H) \) as follows Theorem~\ref{thm:uppernode}.
\qed

\section{COMPUTATIONAL COMPLEXITY}
While both, Theorem~\ref{thm:uppernode} and Theorem~\ref{thm:uppergraph} provide theoretical upper bounds to GNN resilience and the main parameters influencing their computational complexity $ d_{j,i} $ and  $ e_j $ are hence primarily intended as %
analytical constructs, we provide estimates of the computational complexities involved in practically computing $d_{i,j}$ and $ e_j $ below.

To determine time complexity of calculating $ d_{j,i} $ as used in Theorem~\ref{thm:uppernode}, we analyze the expression $ d_{j,i} = \max_{\mathbf{x}_u,\mathbf{x}_v \in {Q}^{n_{j,i}}}\left | \mathbf{x}_u-\mathbf{x}_v \right |_0 $. This represents the maximum $ \ell_0 $ norm of the difference between node embeddings within the space $ {Q}^{n_{j,i}} $. The $ \ell_0 $ norm measures the number of entries where two vectors $ \mathbf{x}_u $ and $ \mathbf{x}_v $ differ. The search space consists of all possible pairs of vectors in $ {Q}^{n_{j,i}} $, yielding $|Q|^{2n_{j,i}}$ pairs. Calculating the $ \ell_0 $ norm for each pair requires iterating over all $ n_{j,i} $ dimensions, resulting in a time complexity of $ \mathcal{O}(n_{j,i}) $ for each pair. Consequently, the overall complexity for evaluating all pairs to find the maximum $ \ell_0 $ norm is $\mathcal{O}(|Q|^{2n_{j,i}} \cdot n_{j,i})$. This reflects the worst-case scenario where every pair of node embeddings is considered. Thus, $ d_{j,i} $ can be calculated in $\mathcal{O}(|Q|^{2n_{j,i}} \cdot n_{j,i})$ time.

To determine the time complexity of calculating both $ d_{j,i} $ and $ e_j $ as used in Theorem~\ref{thm:uppergraph}, we must consider the process of identifying the graph pair $ G_e $ and $ H_e $, which maximize the WL difference $\Delta_{WL}$. The value $ d_{j,i} $ is defined as the maximum $\ell_0$ norm between node embeddings $\mathbf{x}_u$ and $\mathbf{x}_v$ from graphs $ G_e $ and $ H_e $ respectively, specifically $ d_{j,i} = \max_{\mathbf{x}_u, \mathbf{x}_v \in {Q}^{n_{j,i}}} | \mathbf{x}_u - \mathbf{x}_v |_0 $. The process starts by identifying $ G_e $ and $ H_e $ as the pair of graphs in the dataset ${D}$ with equal numbers of vertices that maximize $\Delta_{WL}(G, H, j)$. This involves evaluating all graph pairs, leading to $\mathcal{O}(|D|^2)$ combinations, where $|D|$ is the number of graphs. For each pair, the WL difference calculation requires running the WL algorithm over $ k $ iterations with a complexity of $\mathcal{O}(k \cdot (n + m))$, where $ n $ is the number of vertices and $ m $ is the number of edges. Once $ G_e $ and $ H_e $ are identified, the complexity of calculating $ d_{j,i} $ involves computing the $\ell_0$ norm for each node pair across the two graphs, yielding a complexity of $\mathcal{O}(|V(G_e)|^2 \cdot n_{j,i})$. Thus, the overall complexity for $ d_{j,i} $ is $\mathcal{O}(|D|^2 \cdot k \cdot (n_{max(D)} + m_{max(D)}) + |V(G_e)|^2 \cdot n_{j,i})$ with $ n_{max(D)} $ is the maximal number of vertices and $ m_{max(D)} $ is the maximal number of edges of any $G \in D$. For $ e_j $, the complexity is primarily driven by identifying the pair $ G_e $ and $ H_e $, resulting in $\mathcal{O}(|D|^2 \cdot k \cdot (n_{max(D)} + m_{max(D)}))$, capturing the need to evaluate the WL difference across all graph pairs. These complexities reflect the calculations needed for maximizing graph differentiation and node embedding discrepancies.

\begin{table*}[!ht]
\centering
\caption{Overview of Selected Datasets}
\label{tab:datasets}
\begin{tabular}{lcccccc}
\\
\textbf{DATASET} & \textbf{TYPE} & \textbf{\#GRAPHS} & \textbf{AVG. NODES} & \textbf{AVG. EDGES} & \textbf{LABELS} & \textbf{TASK} \\
\midrule
MUTAG        & Chemical  & 188   & 17.9  & 19.8  & Node, Edge  & Class. \\
AIDS         & Chemical  & 2,000 & 15.7  & 16.2  & Node, Edge  & Class. \\
PTC\_FM      & Chemical  & 349   & 14.1  & 14.5  & Node, Edge  & Class. \\
PTC\_MR      & Chemical  & 344   & 14.3  & 14.7  & Node, Edge  & Class. \\
NCI1         & Chemical  & 4,110 & 29.9  & 32.3  & Node        & Class. \\
PROTEINS     & Protein   & 1,113 & 39.1  & 72.8  & Node        & Class. \\
ENZYMES      & Protein   & 600   & 32.6  & 62.1  & Node        & Class. \\
MSRC\_9      & Image     & 221   & 40.6  & 72.4  & Node        & Class. \\
MSRC\_21C    & Image     & 563   & 77.5  & 142.8 & Node        & Class. \\
IMDB-BINARY  & Social    & 1,000 & 19.8  & 96.5  & -        & Class. \\
\bottomrule
\end{tabular}
\end{table*}

\section{RESILIENCE TO RANDOM BIT FLIPS}
Above formulations on the bounds of the numbers of flips required to degrade graph level and node level expressivity allow us to formulate bounds on the probabilities $P_{node}$ and $P_{graph}$ that a random bit flip will degrade a GNNs node or graph expressivity. In the node level case, where node level expressivity would diminish guaranteed in $\mathcal{O}(\max_{\mathbf{x}_u,\mathbf{x}_v \in {Q}^{n_{k,l}}}\left \| \mathbf{x}_u-\mathbf{x}_v \right \|_0 \cdot m_{k,l} \cdot b)$ bit flips for $u,v \in V_{\cup}({D})$ by a dedicated attacker, for random flips we would have to replace $b$ with $2^b$, covering all possible combinations of flips in the $(k,l)$th layer of the GNN. Note we chose the $(k,l)$th layer here as only in this case a loss of injectivity would guarantee a loss of expressivity.
\begin{equation}
\label{eq:pnode}
    P_{node} \leq \left (n_{k,l} \cdot m_{k,l} \right )^{-\beta_{node}}
\end{equation}
with $\beta_{node} = {\max_{\mathbf{x}_u,\mathbf{x}_v \in {Q}^{n_{k,l}}}\left \| \mathbf{x}_u-\mathbf{x}_v \right \|_0 \cdot m_{k,l} \cdot 2^b}$ for $u,v \in V_{\cup}({D})$.
Likewise, we can arrive at a similar formulation for graph level expressivity
\begin{equation}
\label{eq:pgraph}
    P_{graph} \leq \left (n_{k,l} \cdot m_{k,l} \right )^{-\beta_{graph}}
\end{equation}
with $\beta_{graph} = e_k \cdot d_{k,l} \cdot m_{k,l} \cdot 2^b$ and $e_k = \Delta_{WL}(G_e, H_e, k)$, $d_{k,l} = \max_{\mathbf{x}_u,\mathbf{x}_v \in {Q}^{n_{j,i}}}\left \| \mathbf{x}_u-\mathbf{x}_v \right \|_0$, $u \in V(G_e), v \in V(H_e)$ for $G_e, H_e = \argmax_{G,H \in {D}}\Delta_{WL}(G,H,k)$.
Both equations~\eqref{eq:pnode} and~\eqref{eq:pgraph}, %
can be adjusted to the special cases of ReLU activations, homophily and one hot encoded inputs and general layers (i.e. not only the last layer), leading to lower bounds for the given probabilities.

\section{DATASETS}

We use a diverse set of ten real-world datasets to evaluate the intricate interplay between robustness and expressivity of GNNs, with each dataset chosen for its relevance to specific domains and graph structures. These datasets are part of the well-established TUDataset collection~\cite{morris2020tudataset}, which has become a benchmark standard for graph classification and regression tasks. An overview of the datasets is provided in Table~\ref{tab:datasets}.

\paragraph{Chemical Compounds:} The MUTAG, AIDS, PTC\_FM, PTC\_MR, and NCI1 datasets represent chemical compounds, where graphs model molecules with nodes as atoms and edges as chemical bonds. MUTAG is one of the earliest and most commonly used datasets for graph classification, consisting of nitroaromatic compounds classified based on their mutagenicity on Salmonella typhimurium. Similarly, the AIDS dataset comprises molecular graphs used in anti-HIV drug discovery, with the task of predicting inhibitory activity against HIV. The PTC datasets (FM and MR) include compounds tested for carcinogenicity on rodents, classified by different experimental protocols. NCI1 is a larger dataset derived from the National Cancer Institute’s screening, where the task is to classify compounds based on their activity against non-small cell lung cancer.

\paragraph{Protein Structures:} The PROTEINS and ENZYMES datasets originate from the bioinformatics domain, where the graphs represent protein structures. In PROTEINS, nodes represent secondary structure elements of proteins, such as helices and sheets, connected based on spatial proximity. The classification task involves predicting whether a protein is an enzyme or not. ENZYMES extends this concept by assigning enzymes to one of the six top-level classes of the Enzyme Commission (EC) based on the chemical reactions they catalyze~\cite{borgwardt2005protein, schomburg2002brenda}.

\paragraph{Image Segmentation:} The MSRC\_9 and MSRC\_21C datasets are derived from the MSRC database and are used for semantic image segmentation tasks. Here, graphs are constructed from images where nodes represent superpixels and edges encode the spatial relationships between these superpixels. These datasets challenge GNNs to classify nodes into different categories based on visual content, with MSRC\_21C being an extended version containing more classes and a higher complexity~\cite{neumann2016propagation}.

\paragraph{Social Networks:} The IMDB-BINARY dataset is a social network dataset where each graph represents the collaboration network of actors who have appeared together in movies. The task is to classify these networks based on the genre of the movies, specifically distinguishing between action and romance genres. This dataset exemplifies the challenges posed by real-world social networks, where nodes represent individuals, and edges signify their interactions, requiring the GNN to capture intricate social dynamics~\cite{yanardag2015deep}.

These datasets collectively provide a comprehensive evaluation platform, covering a range of different graph structures and complexities, from small molecular graphs to larger social networks and image-derived graphs. By utilizing such a diverse set of benchmarks, our empirical analysis robustly tests our hypothesis across multiple domains.
\end{document}